# Soccer Player Tracking in Low Quality Video

Eloi Martins

2Ai-School of Technology, IPCA

Barcelos, Portugal

a14855@alunos.ipca.pt

José Henrique Brito

2Ai-School of Technology, IPCA

Barcelos, Portugal

https://orcid.org/0000-0002-4544-4698

*Abstract*—In this paper we propose a system capable of tracking multiple soccer players in different types of video quality. The main goal, in contrast to most state-of-art soccer player tracking systems, is the ability of execute effectively tracking in videos of low-quality. We adapted a state-of-art Multiple Object Tracking to the task. In order to do that adaptation, we created a Detection and a Tracking Dataset for 3 different qualities of video. The results of our system are conclusive of its high performance.

## I. INTRODUCTION (HEADING 1)

Recently some methods to tracking soccer players were developed in accordance of market needs. Being able to perform player tracking creates the possibility of analysing their performance, not only allowing to take their training to a higher level but also identifying their mistakes and the way in which their bodies react during difficult situations that might occur during a match. By identifying and working on individual strengths and weakness, coaches can enhance their players development

The most common methods, used by many soccer teams are GPS-based soccer player tracking systems, where the player has to use a GPS device while playing or training, however the device that must be worn raises some issues, like system intrusiveness, player discomfort, high costs associated to the method, or setup time to name a few.

Given these drawbacks of GPS-based systems, our motivation was to approach the tracking task using Multiple Soccer Player Tracking (MSPT) from the video feed. If we can detect the track of every single player during the matches' video, we will be able of measure the performances and capabilities of every individual without the need to use devices attached to the player's body and also track the players on the opposing team.

On one hand, there are several methods developed to accomplish the task of Multiple Object Tracking (MOT). The approach that we present is the adaptation of a pre-existing method to the specific problem of soccer player tracking. The availability of a Multiple Object Tracking (MOT) [1] led us to adapt it to solve our main goal.

On the other hand, there already commercial products that implement optical player tracking with Artificial Intelligence and Machine Learning, such as the system from SPORTLOGIQ [2], albeit their significant cost. Other systems like Ubitrack [3] use Multiview video to make the tracking task more robust, while obviously further increasing the costs. Some systems like the one from Futbol Drones [4] target small clubs that choose to use soccer video taken from drones.

However, the main challenge of any system designed for people tracking lies in the video image quality, image frame rate and crowed environments, where the false positives and miss matching are common. Generic state of the art people trackers are able to produce good results for videos of pedestrians with a reasonable image quality, both in terms of image resolution and frame rate, of people with reasonably different visual appearance. Specialized trackers for sports videos from commercial systems are also able to track players even though they have similar appearance, but also rely on a reasonable image quality. In our setting, our system is expected to be able to cope with videos with a significant image quality degradation, conditions in which both generic and specialized trackers are likely to fail. The motivation for this is twofold. On one hand we would like to empower small clubs that only have access to single view video of their teams and possibly other teams. On the other hand, we would like the system to be robust to quality degradation introduced by network/internet streaming, where video compression introduces visual artifact that impair the performance of object detectors and trackers. Aggressive compression ratios caused by low transmission bandwidth often cause ordinary people detectors to fail, and which makes tracking infeasible.

Given our motivation, we developed a method based on a MOT for our specific needs. Our method addresses the issue of low-quality video by retraining an existing MOT method on low quality images.

## II. RELATED WORK .

Currently, there are several techniques for tracking multiple soccer players from video. Most of them use Multiview field systems to acquire the data and tackle some of the MOT issues. In [5] and [6] the authors show that using various cameras covering all angles of the field will definitely improve and facilitate the tracking results, but the need to install multiple cameras on the field will raise costs and sometimes it just isn't possible to install the needed additional cameras.

The authors of [7] use a simple approach of adapting a Kalman Filter to tracking multiple soccer players. Kalman filtering is an efficient way to address multitarget tracking. First, they define the state vector for multiple object tracking, then a motion model to determine the soccer player position in next frame is applied. Afterwards their approach defines an observation method for detecting players in the next frame. At the end they measure noise covariance and adjust for it.

Most state-of-art methods follow the tracking-by-detection paradigm that firstly applies a detector to all video frames and obtain detections. Secondly, a tracker is run on the set of detections to perform data association, by assigning the bounding boxes that belong to the same object to a track.

Since the proposal of Faster R-CNN [8] we have had access to a reliable way to do detections with good accuracy and sufficient efficiency. Faster R-CNN combines two networks: region proposal network (RPN) for generating region proposals and a network using these proposals to detect objects. To perform object detection, Faster R-CNN applies the region proposal network to generate a multitude of bounding box proposals for each potential object. Alongside SSD and YOLO, Faster R-CNN is one of the most used object detectors by the computer vision community.

Tracktor++ [9], second place in the MOT Challenge 2019, presents a new concept in the tracking task, where it converts a detector into a tracker (a Tracktor) and combines it with a straightforward re-identification model and a camera motion compensation module.

Tracktor++ pushes the tracking-by-detection approach by using an object detection method to perform all tracking tasks. It basically uses an object detector to reuse the bounding box detected on the previous frame and refine it for the following frame. In addition, it adds two simple extensions to this detector: a re-identification Siamese network and a motion model.

The Detector is the core element of Tracktor. It uses Faster R-CNN with a pretrained ResNet50 backbone. To perform the object detection, Faster R-CNN applies a Region Proposal Network to generate a multitude of bounding box proposals for each protential object. Feature maps for each proposal are extracted via Region of Interest (RoI) pooling [10] and passed to the classification and regression heads.

The Motion model mentioned can be one of two types depending on the camera motion: for sequences with a moving camera, it applies a straightforward camera motion compensation (CMC) by aligning frames via image registration using the Enhanced Correlation Coefficient (ECC) maximization [10]. For sequences with comparatively low frame rates, it applies a constant velocity assumption (CVA) for all objects [11] [12]. The motion model is continuously applied to tracks.

The Re-identification Siamese network (Reid) is used to re-identify objects for which their track was lost, e.g. due to occlusions. Appearance models and Re-identification that use the person features face many problems in crowed situations or where the persons have a similar visual appearance. For example, it´s hard to distinguish players because half of them have the same t-shirt colour and the view isn't close enough to see more features. In [14] the authors showed the importance of leaned reID features for multi-object tracking. This was confirmed by the Tracktor++ authors. The re-identification model is based on appearance vectors generated by a Siamese neural Network [15,16,14]. To that end, it stores killed tracks of a predefined number of frames, compares the distance in an embedding space of each killed track with newly detected tracks and re-identifies tracks via a threshold. The embedding space distance is computed by a Siamese CNN and appearance feature vectors for each of the bounding boxes.

III. PROPOSED METHOD

The goal of the proposed study is to develop a system capable of tracking soccer players in low quality videos of sports events. To achieve the main goal of this study we fundamentally base our system on the adaptation of one of the state-of-the-art pedestrian trackers to our specific situation, Tracktor++.

As a first approximation, we applied the original Tracktor++ to a high-quality soccer video sourced from YouTube, however this led to very unsatisfactory results. The poor performance was due to the fact that the original Tracktor++ implementation is pretrained with the MOT-17 dataset, with vastly different image characteristics, namely in terms of image quality, environment appearance and appearance similarities between the people in the videos. As a consequence, the performance of the object detector and Reid network was impaired, leading to extremely poor results for any evaluation metrics. To address these problems, our approach was to retrain the networks with suitable data, both for the detector and the Reid network, with varying degradation parameters, as detailed below.

A. *Detection dataset creation*

To create our dataset, we used an openly available high-quality (4k) soccer game video that provides a clear image of players, without any zoom changes that difficult the tracking performance.

The new dataset is composed by 3 videos of different views of the soccer field. These videos have approximately 15, 16 and 19 seconds, with 30 FPS, and are therefore composed of 462,497 and 595 frames respectively.

To have optimal detections of the players on each frame, we used the Faster_RCNN_(RESNET50_FPN) pretrained model present in the Detectron2 implementation [17]. Detectron2 is Facebook AI Research´s next generation software system that implements state-of-art object detection algorithms.

The pretrained Faster R-CNN model used from Detectron2 detects several elements on each frame and gives us the bounding boxes coordinates, so we filtered the output to keep objects of the "person" class only, and the detection coordinates were translated into the format needed to retrain the Tracktor Detector model.

To achieve the main goal of our system of dealing with image degradation, we took the 3 videos and degraded them with a h.264 codec with two different Constant Rate Factor (CRF) values, to emulate two types of streams: moderate visual degradation, and aggressive degradation. The CRF values were 40 and 50 respectively. Hence, the videos are the same but with 3 different qualities: Normal (N), h.264 with CRF of 40 (40) and h.264 with CRF of 50 (50). As the detections on normal video quality are extremely good, use them as ground-truth for the training with the lower quality videos.

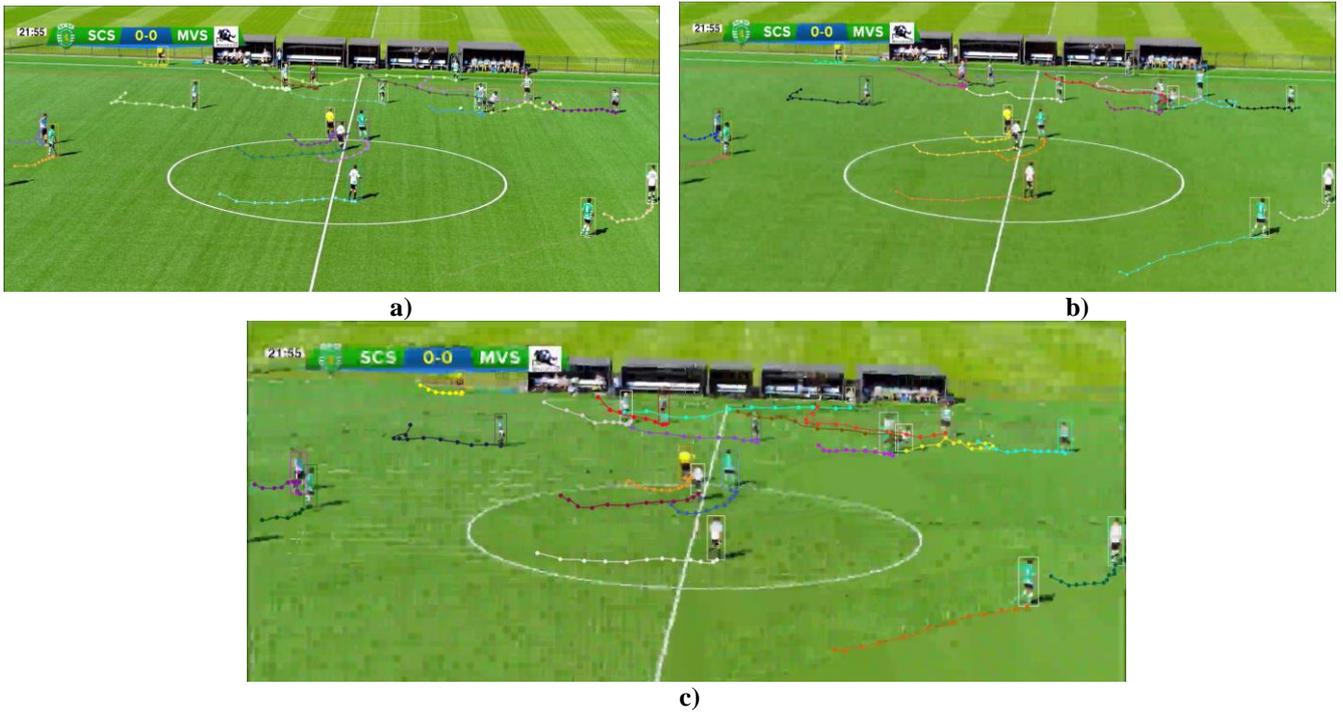

Figure 1. These images are an example of the tracking on the 3 types of degradation. Each dot is the position of the player for every 10 frames: a) tracking results on the normal dataset without degradation; b) tracking results on the CRF 40 degration type dataset c) tracking results on the CRF 50 degration type dataset.

*B. Detector Train*

To reach fair and conclusive metrics we divided the dataset into test and training set. The video sequence with 15sec was chosen as test and the other two videos as training.

We performed our detector training on Google Colab Pro with the best RAM and GPU specifications. The training was executed 3 times, 1 for each video quality (N,40,50), resulting in 3 new detection models that are specific for each type of soccer video stream in which we want to track players

*C. Tracking dataset*

In order to calculate evaluation metrics of our system and the new models, we needed a tracking Ground-Truth of our videos dataset. To do that, we run Tracktor with the N detection model and the corresponding N dataset tracking result from Tracktor++. Tracktor++ outputs a tracking file with the player frame position and ID. We manually rectified the tracks by following each player at a time on the output video sequence.

The tracking ground-truth is the same for every video degradation level, since it is the same sequence just with degraded quality, so we reuse the ground-truth data for the retraining of the Reid network for the 3 types of videos (N,40,50),.

*D. Reid Training*

To retrain Reid, we just needed the tracking ground-truth and the video sequence of our dataset. We didn't know the efficiency of Reid or it is impact on the Tracktor performance, so we retrained a new Reid model for each dataset quality level and evaluated each one.

## IV. EXPERIMENTS AND RESULTS

This section describes the experiments and results regarding the soccer player tracking. The evaluation of our proposed method is through a new dataset of players detections and tracking. All test experiments were implemented on a computer with an Intel Core i7-8700, 32GB of RAM memory and NVIDIA GeForce GTX 1070 WITH 8GGB dedicated memory.

To evaluate our tracker performance, we used Multiple Objects Tracking Accuracy (MOTA) [18] and Multiple Object Tracking Precision (MOTP) metrics. MOTA accounts for all object configuration errors made by the tracker, false positives, misses, mismatches, over all frames. False Positives (FP) that is the number of frames where system results contain at least one object, while ground truth either does not contain any object or none of the ground truth's objects fall within the bounding box of any system object. MOTA quantifies the tracker's ability to follow the objects throughout the image sequence.

MOTP is the total position error for matched object hypothesis pairs over all frames, averaged by the total number of matches made. It shows the ability of the tracker to estimate precise object positions, independent of its skill at recognizing object configurations, keeping consistent trajectories

Tracktor++ has the original Detector model and Reid model that we used as a baseline for our system metrics. Therefore, we have 4 possible Detection models and 4 Reid Models. Tracktor++ combines the Detector network and the Reid network, but it is possible to apply changes only to the Detector, therefore we have 20 possible combinations of tests: each Detector model with each Reid model (4 x 4) and only using the detection network without Reid (more 4 tests). In each combination of Detector and Reid we tested the 3 datasets (N, 40, and 50) and we present the MOTA in Table 1, Table 2 and Table 3 respectively.

**Table 1. MOTA scores for each combination of Detector and Reid using the Normal Dataset**

|  |  | Detector | | | |
|---|---|---|---|---|---|
|  |  | Original | Normal | 40 | 50 |
| REID | Without | 21.4 | 96.3 | 92.1 | 92.1 |
|  | Original | 24.0 | 97.1 | 92.5 | 92.4 |
|  | Normal | 24.0 | 97.1 | 92.6 | 92.5 |
|  | 40 | 24.0 | 97.1 | 92.6 | 92.5 |
|  | 50 | 24.0 | 97.1 | 92.6 | 92.5 |

**Table 2. MOTA scores for each combination of Detector and Reid using the 40 Dataset**

|  |  | Detector | | | |
|---|---|---|---|---|---|
|  |  | Original | Normal | 40 | 50 |
| REID | Without | 22.7 | 90.1 | 92.9 | 93.0 |
|  | Original | 25.2 | 92.0 | 94.1 | 93.7 |
|  | Normal | 25.2 | 91.8 | 94.0 | 93.8 |
|  | 40 | 25.2 | 91.9 | 94.1 | 93.8 |
|  | 50 | 25.3 | 92.0 | 94.1 | 93.8 |

**Table 3. MOTA scores for each combination of Detector and Reid using the 50 Dataset**

|  |  | Detector | | | |
|---|---|---|---|---|---|
|  |  | Original | Normal | 40 | 50 |
| REID | Without | 9.8 | 75.6 | 86.3 | 88.5 |
|  | Original | 12.0 | 79.3 | 88.3 | 90.1 |
|  | Normal | 12.0 | 79.1 | 88.1 | 90.0 |
|  | 40 | 12.0 | 79.2 | 88.2 | 90.1 |
|  | 50 | 12.0 | 79.4 | 88.3 | 90.1 |

At first glance, it is obvious that the newly trained detection models are clearly more capable of detecting soccer players and outperform the original detector by a large margin, as expected.

Beyond this obvious comparison, MOTA scores are notably different between experiment using Reid and experiments ran without Reid. Using it will always improve the tracking performance by ± 2-3%. There are very slight differences between results with the 4 different Reid models for a given detector model. We therefore conclude it makes no significant difference to retrain the Reid model on images for different quality levels, as it does not significantly improve the tracking performance. These MOTA metrics therefore show that the Detector model is the crucial part of our tracking system, as the differences in results between original Detector and the other 3 trained Detectors are significant.

The MOTA results also show that each model combination is the best possible match for the dataset of the degradation type it was trained for, e.g. the Normal Detector model, trained with Normal video, is the best Detector/Reid for the normal test dataset, as expected.

Another significant observation is that every trained model has satisfactory metrics when applied to all dataset. If we apply the Detector trained with degraded images with a CRF of 50, its performance on Normal (high-quality) images is still satisfactory, although obviously not as good as the Normal Detector applied to the Normal dataset. Still, a MOTA of 92.5% is impressive in these conditions.

A final note on the MOTA results, is that the Detector trained on very degraded images (CRF 50, last column of Table 1, Table 2 and Table 3) produces reasonable and fairly stable results on all test datasets, whereas the Detector trained with images with medium degradation (CRF 40, penultimate column of Table 1, Table 2 and Table 3) performs adequately on images with equal or lower degradation then the images it was trained with, by suffers a performance penalty when used in images with a more significant image degradation (Table 3).

In the table 4 we present MOTP results. We expected that MOTP values depended only the detector and the measured results demonstrated our assumption. All the Reid combination with Detector have the same results for MOTP. Due to that, we only present the MOTP of Detector models for each test dataset.

**Table 4. MOTP scores for each combination of Detector and Dataset**

|  |  | Detector | | | |
|---|---|---|---|---|---|
|  |  | Original | Normal | 40 | 50 |
| Data set | Normal | 0.22 | 0.07 | 0.10 | 0.13 |
|  | 40 | 0.24 | 0.15 | 0.13 | 0.13 |
|  | 50 | 0.27 | 0.28 | 0.18 | 0.18 |

Each ground-truth trajectory can be classified as mostly tracked (MT), partially tracked (PT) and mostly lost (ML). This is evaluated based on how much of the trajectory is recovered by the tracking algorithm. If a target is successfully tracked for at least 80% it is classified as MT, if it is between 80% and 20% it is PT, and finally if a track is only recovered for less than 20% it is said to be ML. Good tracking results will be the ones where MT are equal or close to the ground-truth number of tracks. The metrics of

Precision and Recall for each model are also significant. Recall is the number of true positives divided by the sum of the true positives and the false negatives. Recall is the fraction of items that were correctly detected among all the items that should have been detected. Precision is the number of true positives divided by the sum of the true positives and the false positives. That is, precision is the fraction of detected items that are correct.

Following the example of MOTA, we computed MT, PT and ML metrics for all model combinations and we concluded that they have the same numerical results for each Detector model, meaning e.g. the Normal model has the same MT, PT and ML when combined with either of the 4 Reid models or without using Reid. This is because Reid only influences approximately 2-3% of tracking results, so it will only affect a small number of tracks and will not significantly impact the detected track sizes. Hence, we only present MT, PT, and ML values for the different Detectors, each one for the 3 datasets, in Table 4, Table 5 and Table 6. The precision and recall metrics refer only to the Detector performance.

The Ground-truth of the Test Dataset contains 32 Tracks, so that is the value that MT, PT and ML should be compared to. Once again, the Detector trained with CRF50 images behaves reasonably well for all image qualities on the test sets, while the Detector trained with CRF40 images behaves well on image qualities equal or better than the ones it was trained for, but does not work as well in images with CRF50 quality.

**Table 4. MT, PT, ML, Recall and Precision metrics to all the Detectors using N Dataset**

|  | Detector | | | |
|---|---|---|---|---|
|  | Original | Normal | 40 | 50 |
| MT | 2 | 31 | 31 | 30 |
| PT | 5 | 0 | 0 | 1 |
| ML | 25 | 1 | 1 | 1 |
|  |  |  |  |  |
| Recall | 24.6 | 99.3 | 99.4 | 99.3 |
| PRCN | 99.1 | 98.2 | 93.8 | 93.8 |

**Table 5. MT, PT, ML, Recall and Precision metrics to all the Detectors using 40 Dataset**

|  | Detector | | | |
|---|---|---|---|---|
|  | Original | Normal | 40 | 50 |
| MT | 2 | 22 | 26 | 26 |
| PT | 5 | 4 | 4 | 4 |
| ML | 25 | 6 | 2 | 2 |
|  |  |  |  |  |
| Recall | 25.9 | 93.7 | 97.2 | 97.4 |
| PRCN | 98.5 | 98.7 | 97.4 | 96.8 |

**Table 6. MT, PT, ML, Recall and Precision metrics to all the Detectors using 50 Dataset**

|  | Detector | | | |
|---|---|---|---|---|
|  | Original | Normal | 40 | 50 |
| MT | 0 | 17 | 20 | 22 |
| PT | 5 | 6 | 5 | 3 |
| ML | 27 | 9 | 7 | 7 |
|  |  |  |  |  |
| Recall | 13.1 | 82.3 | 91.4 | 93.8 |
| PRCN | 94.5 | 97.5 | 97.4 | 97.4 |

CONCLUSION AND FUTURE WORK

In this paper we have successfully created a system capable of effectively tracking soccer players in low-quality video sequences.

The presented model and methodology have the ability to be easily retrained to new soccer match environments, and other types of image degradation.

For future work, it is intended to extend the Detection and Tracking Datasets for better results in different soccer games. To overcome the occlusions caused and miss IDs we intend to include a method capable of better distinguish the player characteristics.

The developed tracking system may be used a foundation of a more complex system of player performance analysis, by identifying the behavior of a given player, if the player can be reidentified in tracks from multiple videos of different matches.

ACKNOWLEDGMENT

We would like to thank the Applied Artificial Intelligent Laboratory (2AI) for providing access to all conditions for the development of the project.